\documentclass[a4paper,twoside]{article}

\usepackage{epsfig}
\usepackage{subcaption}
\usepackage{calc}
\usepackage{amssymb}
\usepackage{amstext}
\usepackage{amsmath}
\usepackage{amsthm}
\usepackage{multicol}
\usepackage{pslatex}
\usepackage{apalike}
\usepackage{algorithm2e}
\usepackage[bottom]{footmisc}

\usepackage{amssymb}
\usepackage{amsmath}
\usepackage{graphicx}
\usepackage[T1]{fontenc}
\usepackage{url}

\usepackage{SCITEPRESS}     %

\begin{document}

\title{Animating NeRFs from Texture Space: A Framework for Pose-Dependent Rendering of Human Performances}

\author{\authorname{Paul Knoll\sup{1,3}\orcidAuthor{0009-0008-2990-1968}, Wieland Morgenstern\sup{1}\orcidAuthor{0000-0001-5817-7464}, Anna Hilsmann\sup{1,2}\orcidAuthor{0000-0002-2086-0951} and Peter Eisert\sup{1,2}\orcidAuthor{0000-0001-8378-4805}}
\affiliation{\sup{1}Fraunhofer Heinrich Hertz Institute, HHI}
\affiliation{\sup{2}Humboldt University of Berlin}
\affiliation{\sup{3}Johannes Kepler University Linz}
\email{first.last@hhi.fraunhofer.de}
}

\keywords{Computer vision representations, Animation, Rendering}

\abstract{Creating high-quality controllable 3D human models from multi-view RGB videos poses a significant challenge. Neural radiance fields (NeRFs) have demonstrated remarkable quality in reconstructing and free-viewpoint rendering of static as well as dynamic scenes. The extension to a controllable synthesis of dynamic human performances poses an exciting research question. In this paper, we introduce a novel NeRF-based framework for pose-dependent rendering of human performances. In our approach, the radiance field is warped around an SMPL body mesh, thereby creating a new surface-aligned representation. Our representation can be animated through skeletal joint parameters that are provided to the NeRF in addition to the viewpoint for pose dependent appearances. To achieve this, our representation includes the corresponding 2D UV coordinates on the mesh texture map and the distance between the query point and the mesh. To enable efficient learning despite mapping ambiguities and random visual variations, we introduce a novel remapping process that refines the mapped coordinates. Experiments demonstrate that our approach results in high-quality renderings for novel-view and novel-pose synthesis.}

\onecolumn \maketitle \normalsize \setcounter{footnote}{0} \vfill

\begin{figure}[!h]
  \centering
  \includegraphics[width=\textwidth]{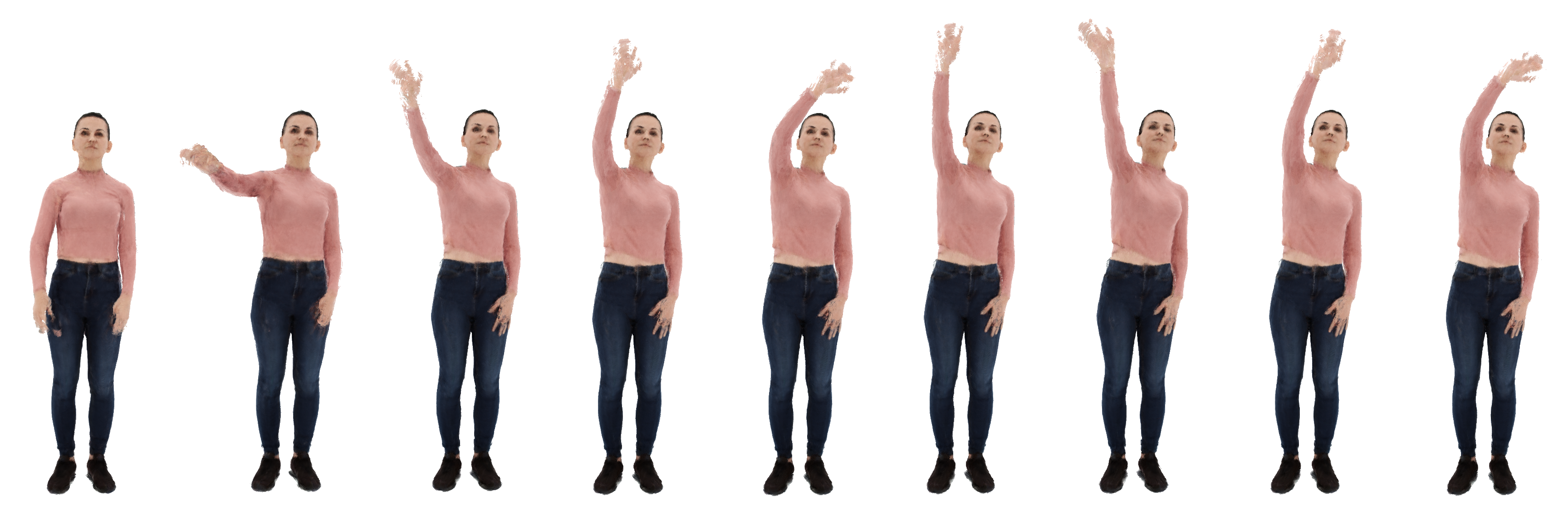}
  \begin{minipage}{\textwidth}
    \centering
    \caption{Synthesized images of the hand wave sequence in novel view and novel pose.}
    \label{fig:teaser}
  \end{minipage}
\end{figure}

\clearpage

\section{Introduction} \label{sec:intro}
Creating fully animatable, high quality human avatars has been a longstanding research question with significant potential impact. In visual applications involving human models, one desirable and essential capability is to render images from different viewpoints while also having the flexibility to manipulate the pose as needed. Traditional methods for generating high-fidelity human avatars often come with a high cost, as they often require manual modelling. Recently, research has shifted towards using only 2D images to capture and represent 3D human models. To achieve this, the solution has to aggregate and distinguish visual information over time as the body performs 3D motion.

Neural Radiance Fields (NeRFs) \cite{nerf} have emerged as a powerful approach for synthesizing photorealistic images of complex static scenes and objects. This technique has shown remarkable success in various domains, such as computer graphics and virtual reality. However, the synthesis of humans in unseen poses with NeRFs has remained a challenging task, primarily due to the high complexity and variability of human appearance. Various methods have been proposed to extend the NeRF framework to encompass novel poses of dynamic humans \cite{2021narf,liu2021neural,animatablenerf,sanerf,zheng2022structured}, but so far none has achieved the fidelity of static NeRFs. The ability to generate controllable human synthesis with NeRFs has the potential to substantially impact a wide range of media production applications, including avatar creation, telepresence, movie production, gaming, and more.

In this paper, we present a novel NeRF-based model, addressing the challenge of synthesizing humans in novel poses and views. Our work is inspired by the idea of Xu et al.\cite{sanerf}, transforming the NeRF query space from absolute world coordinates to surface-aligned coordinates. We find a novel, distinct, more compact representation of the NeRF space around the body, allowing a coherent representation across different poses. Additionally, we introduce a neural network-based remapping of the query coordinates to account for pose-dependent variations.

 Specifically, our contributions are the following:
\begin{itemize}
  \item We extend the NeRF framework to encompass varying poses by introducing a novel surface aligned $uvh$ representation that is persistent between different frames and poses
  \item We introduce a novel $uvh$ remapping process that further refines this novel representation by accounting for body model and fitting inaccuracies
  \item We combine our method with the Instant-NGP \cite{instant-ngp} multiresolution hash encoding and adapt it to deal with our warped coordinate space for significantly improved training times
\end{itemize}

\begin{figure*}[htb]
  \centering
   \includegraphics[width=0.7\linewidth, page=6
   ]{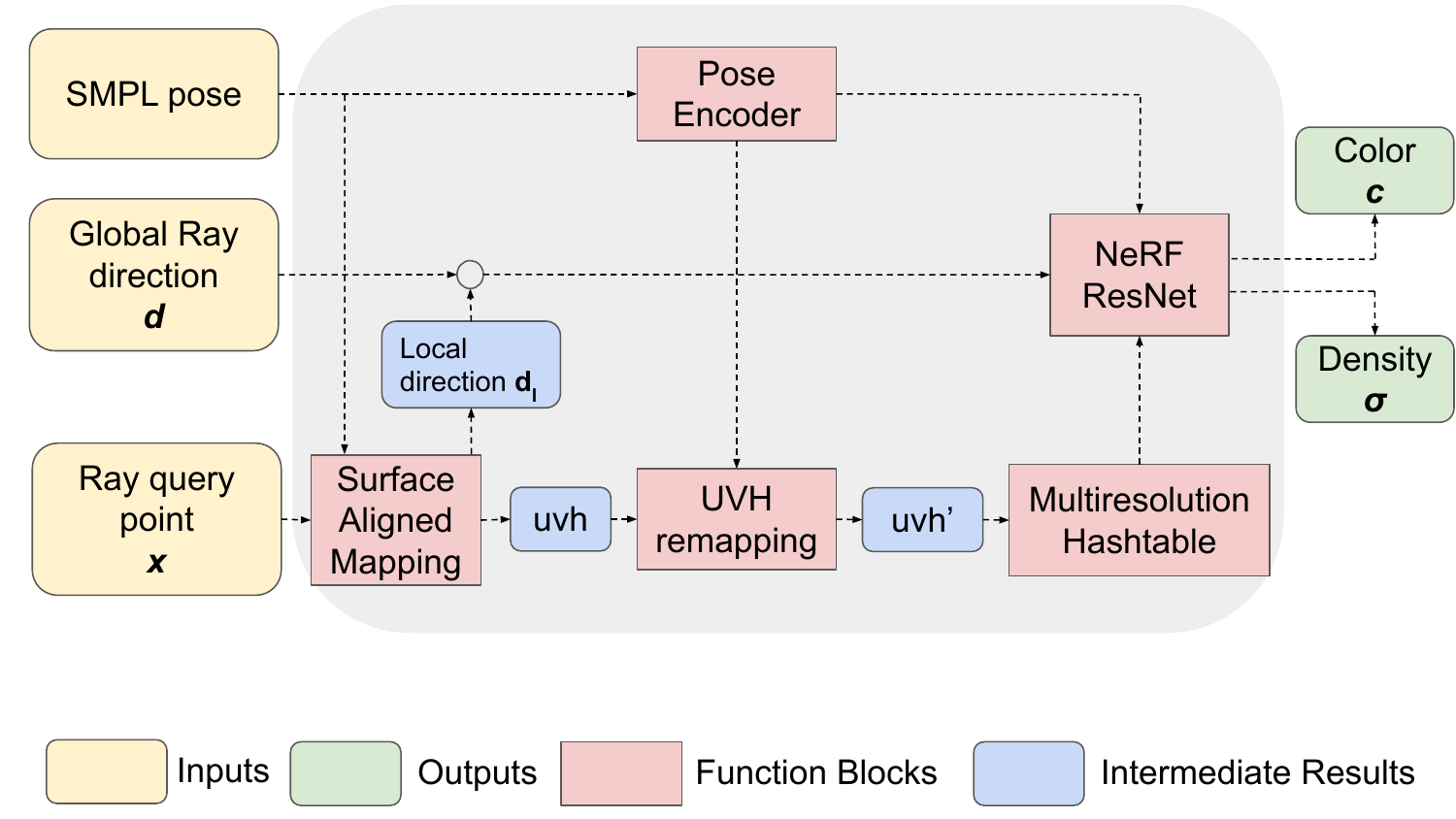}

   \caption{Network architecture of the proposed UVH-NeRF.}
   \label{fig:network_arc}
\end{figure*}

\section{Related Work}

Recent works have extended the Neural Radiance Fields framework to model the time domain of dynamic scenes, including articulated models such as humans in motion \cite{du2021nerflow,nerfies,hypernerf}. 

These approaches can be classified into deformation-based methods such as Nerfies \cite{nerfies} and HyperNeRF \cite{hypernerf}, which mainly focus on the face, and other methods that use human-prior information like skeletons, parametric models like SMPL \cite{smpl}, to account for deformations, such as HumanNeRF \cite{humannerf}. Still, these methods mainly focus on free view-point rendering of the dynamic scenes without the capability of modifying and animating poses.

Very recently, researchers have proposed first approaches towards controllable NeRFs by combining them with mesh-based deformation, e.g.~the SMPL model \cite{animatablenerf,neuralbody,sanerf}. These methods use the SMPL model to learn a deformation field \cite{liu2021neural,2021narf,animatablenerf} or directly learn a NeRF on the surface \cite{sanerf,zheng2022structured}. 

Neural Body \cite{neuralbody} is a mesh-based method that uses a per-vertex latent code to generate a continuous latent code volume. Structured Local Radiance Fields \cite{zheng2022structured} present a distinct approach in which a set of local radiance fields is anchored to pre-defined nodes of a body template. Surface-Aligned NeRFs \cite{sanerf} present another mesh-based approach, where the neural scene representation is defined on the surface points of a human body mesh along with their signed distances from the surface. While our method is inspired by the work of Xu et al.~\cite{sanerf}, we propose a more compact representation of the NeRF space around the body, allowing a consistent representation in pose space. Despite recent advancements in the dynamic rendering of human models using NeRFs with controllable pose, limitations persist, primarily in the form of inferior image quality when compared to static NeRFs.

One challenge with the vanilla NeRF approach and the extensions is the time-consuming nature of the training. To address the multi-day training time, recent works have proposed various encoding schemes to reduce costly backpropagation through deep networks. For instance, Instant-NGP \cite{instant-ngp} employs a learned parametric multiresolution hash encoding, simultaneously trained with the NeRF MLPs. This leads to a significant improvement in training and inference speed, as well as scene reconstruction accuracy. TensorRF \cite{tensorrf}, on the other hand, uses a neural-network-free volume rendering approach, where a scalar density and a vector color feature are stored in a 3D voxel grid. We leverage the training time improvement that Instant-NGP achieved, by incorporating the multi resolution hash encoding in our model.

\section{UVH-NeRF for Pose Control}

Our novel framework for NeRF based human representation with pose control leverages the SMPL \cite{smpl} model as a prior for human poses. Around the model, we construct a novel NeRF space that is based on the projected 2D $uv$ texture coordinates of the SMPL mesh and its distance to the ray query point. Thereby, we obtain a novel space (in which we embed the NeRF ray query coordinates) that is fixed for all frames and poses. In addition, the NeRF is controlled by a set of encoded joint angles representing body pose. Though this procedure, we achieve a fully animatable human model that can be animated by simply adapting the SMPL pose parameters to the desired pose, while the NeRF is queried to render high-quality free-viewpoint pose-dependent images.
In the following sections, we explain the utilized data and the architecture of our model in detail.

\subsection{Datasets}
We use two different datasets for training and evaluation. 

First, we use our own dataset consisting of recordings of an actress performing a variety of tasks. These recordings consist of 16 equally spaced camera pairs on three different height levels. The images are captured with a resolution of 3840x5120 pixels and downscaled to 960x1280 pixels for further processing. They show the actress in a well-lit environment in front of a bright background. We asked the actress to perform a variety of tasks, including hand gestures, talking and whole body gestures, to aid us in building an interactive model. The selection of the frames used for training is described in Section \ref{experiments}.

Additionally, we use the ZJU-MoCap dataset \cite{zjumocap,neuralbody} from Zhejiang University, which was recorded using 21 cameras on the same height level around an actor performing natural movements, for evaluation. The ZJU data already contains fitted SMPL parameters.

\subsubsection{Preprocessing: SMPL Pose and Shape Fitting} \label{sec:smpl_fitting}
\begin{figure}[htb]
  \centering
   \includegraphics[width=1\linewidth]{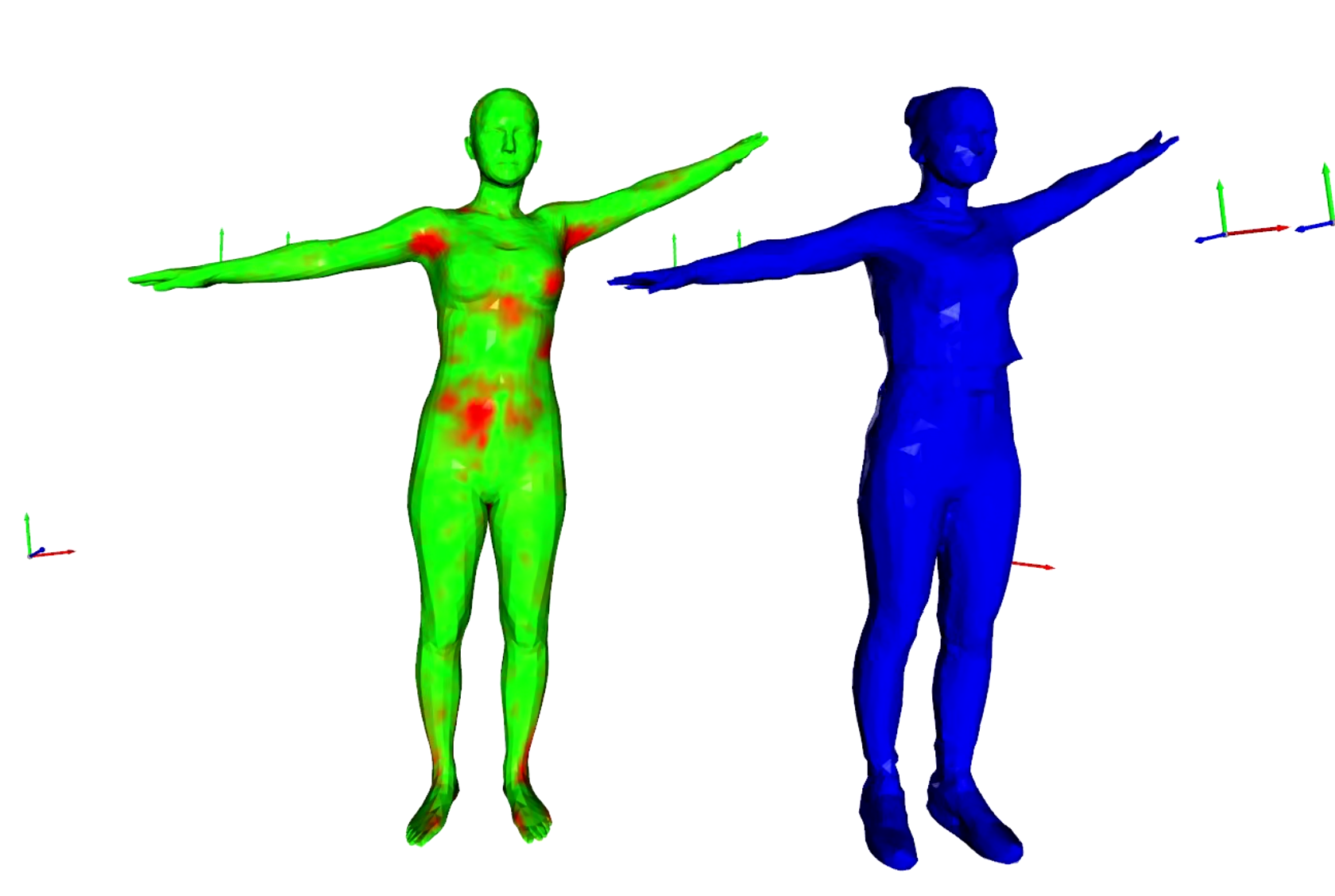}

   \caption{Depiction of complete SMPL fitting process. Left: fitted SMPL model with distance to scan coloured in green and red. Green represents a distance of zero, red larger distances (up to a threshold). Right: scan, shifted to the right for better comparison. The arrows in the background represent the locations of the cameras.}
   \label{fig:sarah_mesh_opt}
\end{figure}

The suggested NeRF framework leverages the SMPL model as a prior for human body shape and pose and thus needs the corresponding pose and shape parameters for each frame of the training sequence. To estimate SMPL parameters from recorded sequences, we use an improved version of EasyMocap \cite{easymocap}.
EasyMocap is an open source toolbox for human pose and shape estimation from multi-view videos. 
While the original version of EasyMocap relies on OpenPose \cite{openpose} keypoints for the optimization, we expanded it to incorporate an additional loss on the 3D mesh reconstructed from the multiview data \cite{decai22} (which we will call \textit{scan} in the following). 
For this loss, we calculate the shortest distance between the vertices of the current best SMPL estimate, and the faces of the scan during optimization of the shape and body pose parameters. 
An example of a fitted frame is depicted in Figure \ref{fig:sarah_mesh_opt}.

\subsection{Network Overview}

In this section, we present the architecture of our proposed model. Our goal is to facilitate rendering of human representations in unseen poses. In order to achieve this objective, we have integrated concepts from coordinate transformation, encoding and remapping, along with a preprocessing of the query points. With these extensions, our method differs significantly from the original NeRF network \cite{nerf}: The changes made to the original model primarily focus on the processing of 3D query points to extract density and color values. While our method provides a novel parametrized scene representation, we can use off-the-shelf NeRF sampling methods when querying our customized scene to form the final 2D image. We choose to use the ray marching procedure for sampling and combining points into an image from \cite{instant-ngp} and will not further explore the topic of sampling in this section.

Firstly, in order to allow for pose-dependent rendering, we transform the set of query points to a novel surface-aligned coordinate system, warped around the mesh surface, utilizing the SMPL model as a prior. During training, the query points in the new surface-aligned space are remapped using a learned frame embedding, correcting inaccuracies arising from both the body prior and the fitting process. Finally, the new coordinates are processed in a modified NeRF network together with the encoded body pose.
The network's primary workflow, function blocks, input, and output parameters are illustrated in Figure \ref{fig:network_arc}. They will be explained step-by-step in the following paragraphs.

Additionally to the ray direction $\mathbf{d} \in \mathbb{R}^3$ and the ray query point $\mathbf{x} \in \mathbb{R}^3$, which serve as inputs for the standard NeRF model, we provide the SMPL pose and shape parameters of the corresponding frame to the network to learn a pose-dependent representation of color and density. The SMPL parameters are used in various ways in this approach, and are estimated in advance, as detailed in Chapter \ref{sec:smpl_fitting}. 

The initial estimations of the pose parameters are then further improved during the NeRF training, by leveraging a differentiable SMPL model, while the body shape estimations stay constant for all frames of the same person. The resulting improvement in rendering quality is discussed in Chapter \ref{experiments}.

\subsubsection{Surface-Aligned Mapping}
To enable the model to represent humans in different poses and motion, the coordinate system we operate in is attached to a controllable surface mesh
instead of using a static global one. The goal is to obtain the density and color values of a specific point in the global coordinate system from a nearby point on a surface mesh and its respective distance. This way, when the pose of the human changes, the body prior can be modified accordingly with the corresponding parameters, and the network is still able to reconstruct the scene correctly. Generally, this approach allows for a simple way to model dynamic objects of all kinds given a deformable mesh. In this work, the human body is implemented using SMPL \cite{smpl}.  

The surface aligned function block is denoted by $M_{sa}(\mathbf{x}, \vec{\beta}, \vec{\theta})$, with SMPL shape parameters $\vec{\beta} \in \mathbb{R}^{10}$, SMPL pose parameters $\vec{\theta} \in \mathbb{R}^{72}$, and 3D query point $\mathbf{x}$.
The frame specific surface mesh is reconstructed with the pre-fitted pose and shape parameters.
For each query point $\mathbf{x}$, a corresponding surface point $\mathbf{s}$ on the mesh is found by the dispersed projection method, as detailed in \cite{sanerf}. We use the dispersed projection method instead of the nearest point projection, as it does not suffer from the indistinguishability issue near the edges of the mesh. 
Thereby, new surface aligned query points $uvh \in \mathbb{R}^3$ are obtained, consisting of the $uv$ texture coordinates of the corresponding surface point $\mathbf{s}$ and the signed distance $h$ from the query point $\mathbf{x}$ to $\mathbf{s}$.  While $u$ and $v$ are naturally in $[0, 1]$, the height is normalized by the maximum considered distance $h_{max}$ from the mesh surface. If the query point has a distance of more than $h_{max}$ to the mesh surface, a density of zero is returned.

Furthermore, a new local view direction $\mathbf{d_l}$ is produced by the function block.  
The local view direction $\mathbf{d_l}$ is defined in the local coordinates, i.e. the coordinates formed by the normal, tangent, and bi-tangent directions on the surface of the posed mesh at the projected point $\mathbf{s}$ \cite{sanerf}.
In contrast, the global view direction $\mathbf{d}$ is defined in the global coordinate system, and represents the global direction of the ray.

The local and global view direction are encoded using a concatenation of the original (3D) values and the frequency encoding \cite{vaswani2017attention} function shown in Equation (\ref{eq:freq_enc}), scaling the input according to $L$ frequency bands and injecting the result into alternating sine and cosine functions:

\begin{multline} \label{eq:freq_enc}
    \gamma(p) = \bigl( \sin(2^0 \pi p), \cos(2^0 \pi p), ..., \\
    \sin(2^{L-1} \pi p), \cos(2^{L-1} \pi p) \bigr)
\end{multline}  

where $L = 6$, leading to encoded directions $\mathbf{d}, \mathbf{d_l} \in \mathbb{R}^{39}$.
By transforming the input into a higher-dimensional representation, this encoding method enables easier pattern recognition within the input data, potentially improving the model's ability to learn and generalize.
The encoded global and local view directions are concatenated and used together in the NeRF ResNet color subnetwork, similarly to the standard NeRF procedure \cite{nerf}.

While the SMPL shape is only used directly for the mesh generation, in the \textit{surface aligned mapping} function block, the pose is encoded in the \textit{pose encoder} $E_{pose}: \mathbb{R}^{78} \mapsto \mathbb{R}^{P_L}$ for later use. We adapt the latent pose dimension $P_L$ with respect to the number of different poses the model had to account for. The pose encoder $E_{pose}$ consists of a simple neural network with one linear layer and a hyperbolic tangent (Tanh) output activation function and is trained jointly with the rest of the network.
$E_{pose}$ is employed to create a rich and informative latent code of the pose dependent features and differences in appearance. 

\subsubsection{UVH Remapping}
As the surface geometry of the SMPL model cannot model the real human surfaces in sufficient detail and due to inaccuracies and variations in fitting the model to real data,
we introduce a uvh-remapping subnetwork to enable the network to learn a sharp representation despite these misalignments.
The network takes two inputs: the learned pose embedding $\vec\theta_{enc}$, and the surface aligned coordinates $uvh$. The network calculates an offset $\Delta uvh = F_{\psi}(uvh, E_{pose})$, where $\psi$ represents the weights of the neural network. From these offsets, we obtain the remapped surface aligned coordinates $uvh' = uvh + \Delta uvh$.

The neural network $F_{\psi}$ of the remapping module consists of three linear layers with ReLU activation and a scaled Tanh output activation function. This way, the maximum possible distance the network can remap the query points is limited, to avoid undesired large offsets. Furthermore, an additional loss is introduced by the sum of the total offsets, to keep them limited to where they are really useful and necessary for a sharp representation.

\subsubsection{Multiresolution Hash Encoding}
In the original NeRF paper, the authors found that having the NeRF network $F_\Theta$ operate directly on the 3D query points $\mathbf{x}$ leads to poorly represented high frequency variation in color and geometry \cite{nerf}. Therefore, the authors employ the frequency encoding in Equation (\ref{eq:freq_enc}) to map the inputs to a higher dimensional space. Subsequently, in the Instant-NGP paper \cite{instant-ngp}, a new improved encoding method named multiresolution hash encoding was proposed, which significantly speeds up training time. %

The technical details of this approach can be found in \cite{instant-ngp}.
Fundamentally, voxel grids on different resolution levels are employed, containing learnable parameters in their corners. While the parameters can be mapped one to one to the corners on the coarser levels, on the finer levels a hash mapping is introduced. The values of the corresponding corner parameters on each level are then interpolated linearly and concatenated to obtain a representation for every resolution. While Instant-NGP operates in the static 3D space, our proposed model resides in the warped space around the mesh of the given person.
In our approach, we feed the remapped surface aligned coordinates $uvh'$ into the multiresolution hash encoding $E_{hash}$ instead of the absolute 3D world coordinates $\mathbf{x}$ in Instant-NGP. This leads to $E_{hash}: \mathbb{R}^{N \times 3} \mapsto \mathbb{R}^{N \times (L \cdot F)}$, with the number of levels $L = 16$ and feature dimension $F = 4$. The remaining parameters of the encoding stay the same as in \cite{instant-ngp}, while $N$ represents the dynamic number of query points in one forward pass of the network.

\subsubsection{NeRF ResNet}
In order to enable the network to account for various pose dependent feature variations in the same surface aligned coordinate $uvh'$, we introduce a NeRF ResNet into the architecture. The idea of a NeRF ResNet was introduced by Xu et al.~\cite{sanerf}. Its architecture is similar to the one of the original NeRF network \cite{nerf}, which can be divided into two separate MLPs. The first one takes the encoded query point $\textbf{x}_{enc}$ and produces the density $\sigma$ and a feature vector. The second one takes the feature vector and the encoded ray direction $\textbf{d}_{enc}$ and produces the RGB color values.
While we maintain the second network unchanged, the first one is sustituted by a ResNet \cite{resnet}, consisting of 5 residual blocks with a layer width of 64 neurons., with the encoded pose $\vec\theta_{enc}$ as an additional (residual) input.

The NeRF ResNet receives the encoded remapped surface aligned coordinates $uvh'_{enc}$, and the encoded pose $\vec\theta_{enc}$ as inputs, 
producing the density $\sigma$ and the feature vector. 
Finally, the feature vector and the encoded and concatenated local and global view direction $d_{enc} \oplus d_{l,enc}$, are fed into the RGB network, consisting of 2 linear layers, ReLU activation function and Sigmoid output activation, producing the final RGB values.

With this, our new model is complete. Thanks to the incorporation of the ResNet and the remapping module, the network has two mechanisms to account for pose dependent visual variations in our new surface aligned space that would otherwise result in blurry representations. In essence, the remapping is intended to correct the imperfections of the body model and its fitting by mapping the $uvh$ coordinates to their ideal position, while the ResNet  is capable of accommodating pose-dependent differences in appearance, such as clothing folds. As a result, we obtain a fully animatable human body model that can be controlled by adapting the SMPL pose parameters $\vec\theta$ to the desired values.

\section{Experiments} \label{experiments}

To demonstrate the effectiveness of our model, we implemented it on top of the torch-ngp \cite{torch-ngp} library, which is a reimplementation of Instant-NGP \cite{instant-ngp} leveraging the expressiveness of Python and PyTorch \cite{pytorch}.

We conducted different experiments and provide an additional ablation study. In all experiments, we left one camera out to be able to evaluate the quality of the novel view synthesis, and used the remaining ones for training. 

To mitigate the influence of the background in the evaluation, the Peak Signal-to-Noise Ratio (PSNR) is calculated only on the non-masked area of the ground truth images, covering only the human body. This results in reduced PSNR values compared to other approachesCreating high-quality controllable 3D human models from multi-view RGB videos poses a significant chal-
lenge. Neural radiance fields (NeRFs) have demonstrated remarkable quality in reconstructing and free-
viewpoint rendering of static as well as dynamic scenes. The extension to a controllable synthesis of dynamic
human performances poses an exciting research question. In this paper, we introduce a novel NeRF-based
framework for pose-dependent rendering of human performances. In our approach, the radiance field is warped
around an SMPL body mesh, thereby creating a new surface-aligned representation. Our representation can be
animated through skeletal joint parameters that are provided to the NeRF in addition to the viewpoint for pose
dependent appearances. To achieve this, our representation includes the corresponding 2D UV coordinates
on the mesh texture map and the distance between the query point and the mesh. To enable efficient learning
despite mapping ambiguities and random visual variations, we introduce a novel remapping process that re-
fines the mapped coordinates. Experiments demonstrate that our approach results in high-quality renderings
for novel-view and novel-pose synthesis. that calculate the PSNR on the whole image, a methodological difference that does not diminish the satisfactory visual quality achieved by our model.

We train all components of our model simultaneously. We train the models for 100,000 steps with a batch size of 1 on the dataset. This results in training times of less than 12 hours on a GeForce RTX 3090. The rendering of novel images in 1024x1024 resolution takes about 25 seconds in our model, which is not optimized for speed. We leverage the Adam optimizer \cite{kingma2014adam} with a decaying learning rate from 1e-2 to 1e-3.

To demonstrate the expressiveness of our model, we train it on 20 equally spaced frames of our arm rotation sequence. In this sequence, the actress stands still with one arm to the side while the other arm is performing a full rotation. We selected this sequence to show that even this very limited training pose space enables the model to generalize to unseen poses.

\begin{figure}[tb]%
    \centering
    \subfloat[\centering]{{\includegraphics[width=0.5\linewidth]{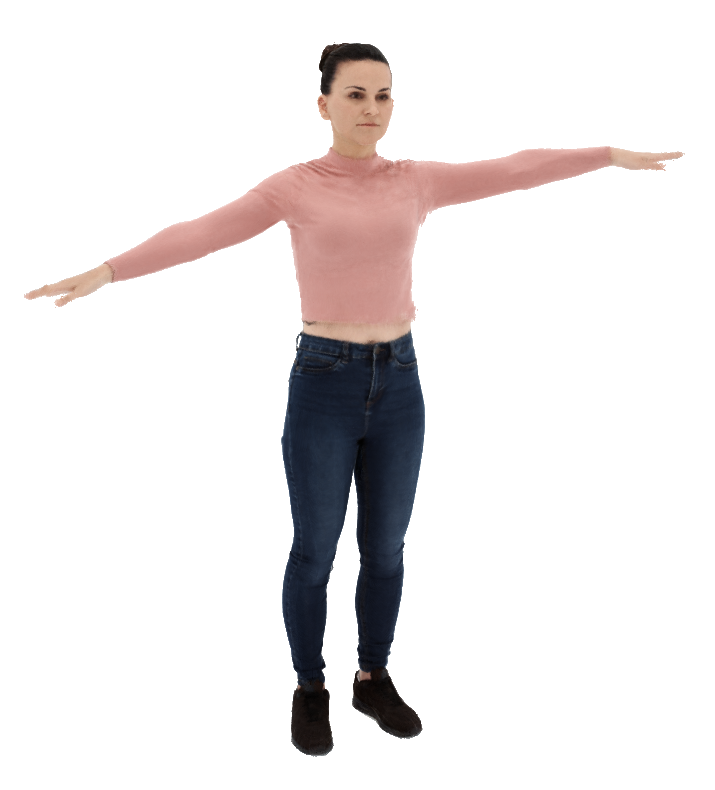} }}%
    \subfloat[\centering]{{\includegraphics[width=0.5\linewidth]{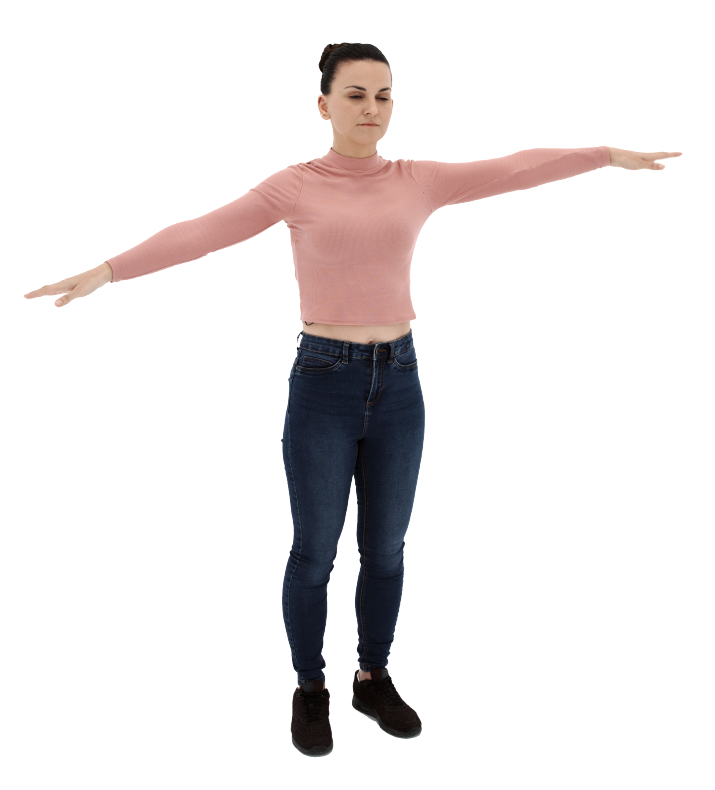} }}%
    \caption{Qualitative comparison of (a) novel view synthesis and (b) ground truth}%
    \label{fig:sarah_novel_view}%
\end{figure}

\begin{figure}[tb]%
    \centering
    \subfloat[\centering]{{\includegraphics[width=0.5\linewidth]{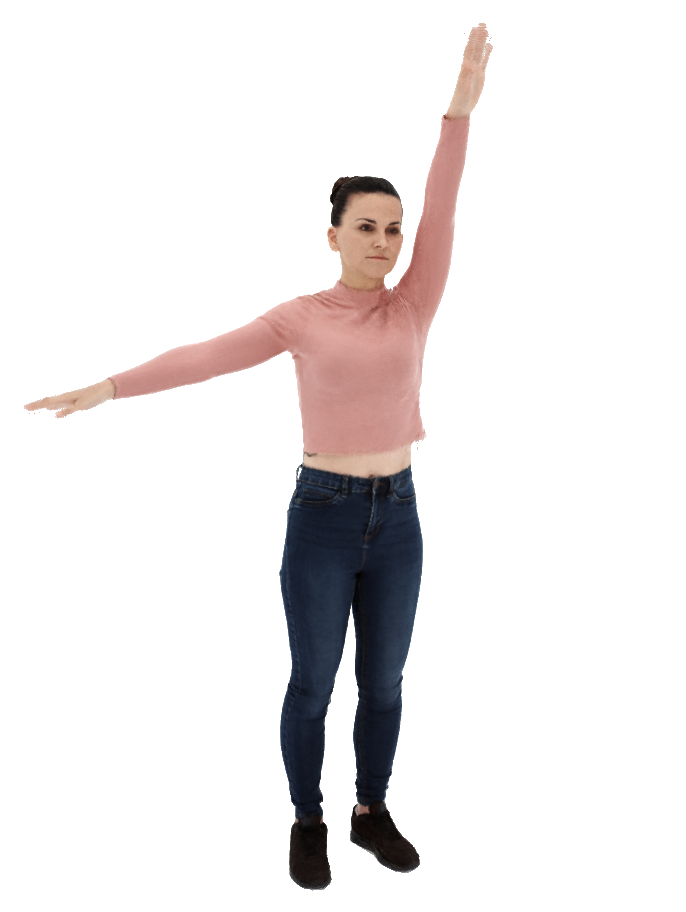} }}%
    \subfloat[\centering]{{\includegraphics[width=0.5\linewidth]{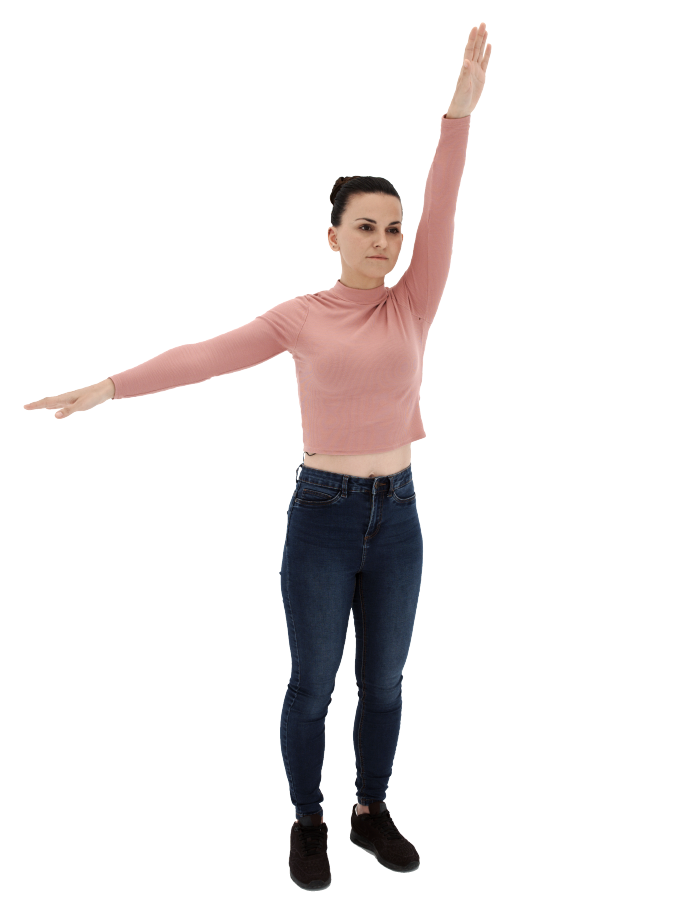} }}%
    \caption{Qualitative comparison of (a) novel view synthesis and (b) ground truth}%
    \label{fig:sarah_novel_view2}%
\end{figure}

Figure \ref{fig:sarah_novel_view} and \ref{fig:sarah_novel_view2} show exemplary qualitative results of the novel view synthesis of this model. The model is able to accurately reconstruct the image while producing a natural and visually pleasing appearance. In Figure \ref{fig:sarah_novel_view}, the rendered subject has her eyes open, contrasting the closed eyes in the ground truth image. This discrepancy illustrates the model's capacity to infer visual features, such as the eyes, from other training frames. Figure \ref{fig:sarah_novel_view2} demonstrates accurate reconstruction of pose depended features in relation to Figure \ref{fig:sarah_novel_view}, e.g. in the lifted shirt and the folds and shadows of the shirt of the actress as her arm rises. On the 20 images of this validation set, we achieved an average PSNR of 28.64 dB.

\begin{figure}[tb]%
    \centering
    \subfloat[\centering]{{\includegraphics[width=0.45\linewidth]{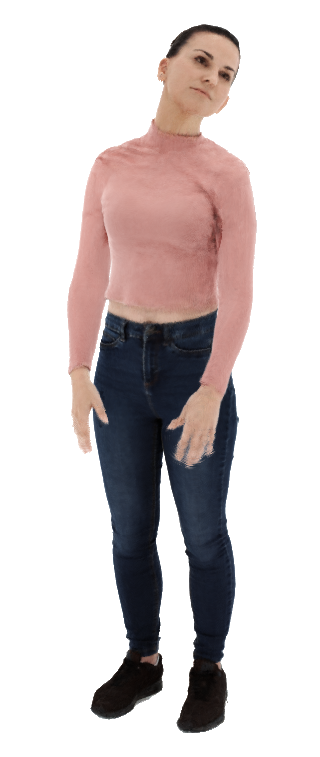} }}%
    \subfloat[\centering]{{\includegraphics[width=0.45\linewidth]{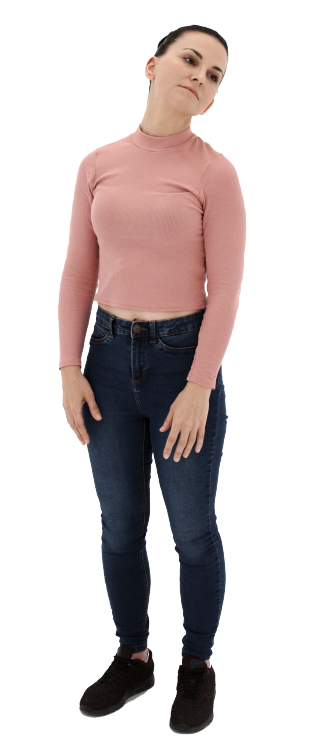} }}%
    \caption{Qualitative comparison of (a) novel pose synthesis and (b) ground truth.}%
    \label{fig:sarah_novel_pose}%
\end{figure}

\begin{figure}[tb]%
    \centering
    \subfloat[\centering]{{\includegraphics[width=0.45\linewidth]{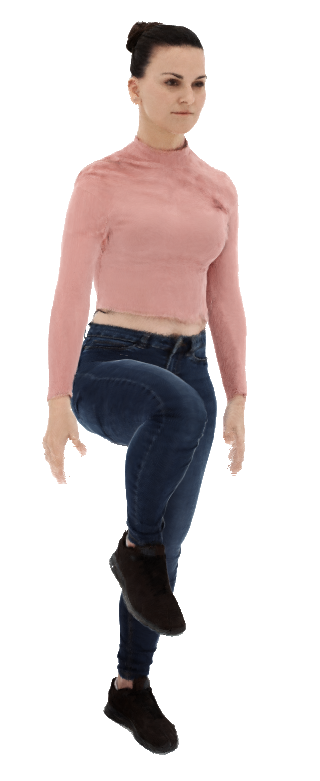} }}%
    \subfloat[\centering]{{\includegraphics[width=0.45\linewidth]{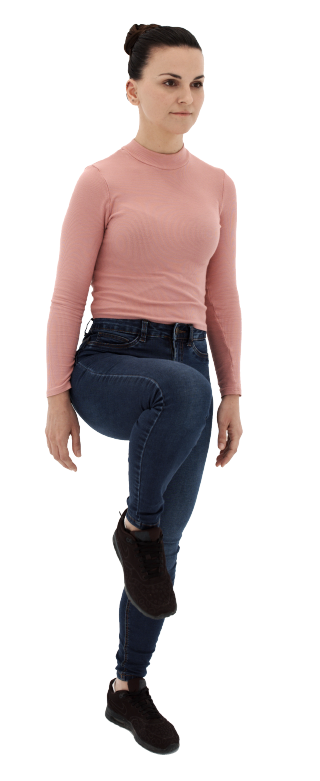} }}%
    \caption{Qualitative comparison of (a) novel pose synthesis and (b) ground truth.}%
    \label{fig:sarah_novel_pose2}%
\end{figure}

We also test this model on novel poses, which is the primary focus of our work. For this, we identify a set of frames with highly diverse poses by performing a cluster analysis on the SMPL parameters of each frame of our datasets. We then manually selected 10 frames from different pose clusters that show natural poses.
By testing those frames with all of our training cameras, we achieved an average PSNR of 19.16 dB for the novel pose synthesis. 
Figure \ref{fig:sarah_novel_pose} and  \ref{fig:sarah_novel_pose2} show qualitative results of novel pose synthesis. These two poses were selected to show the ability of the network to produce natural high fidelity results of poses completely distinct to the training poses. Figure \ref{fig:sarah_novel_pose} shows the actress with a tilted head, a pose that was not seen during training.
Figure \ref{fig:sarah_novel_pose2} shows a synthesized pose where the actress lifts one leg up, while she was standing on both legs in all training frames. While the result is visually plausible, the PSNR decreases. This is expectable as (i) the model adheres to the leg positioning of the SMPL model and encounters challenges to account for pose dependent variations, thus, a slight misalignment between the synthesized leg and the corresponding location in the ground truth is observed, and (ii) remains of the folds around the neck of the actress are visible in the synthesized images, as the network has not seen images with both arms in a lowered position. 

\begin{figure}[tb]%
    \centering
    \subfloat[\centering]{{\includegraphics[width=0.45\linewidth]{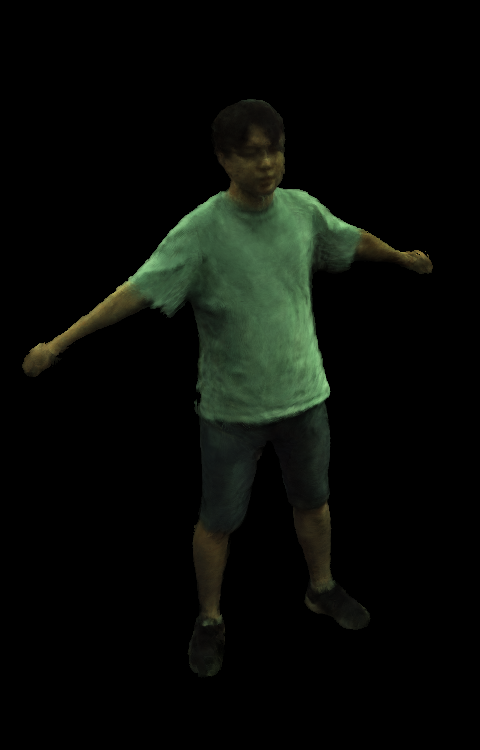} }}%
    \subfloat[\centering]{{\includegraphics[width=0.45\linewidth]{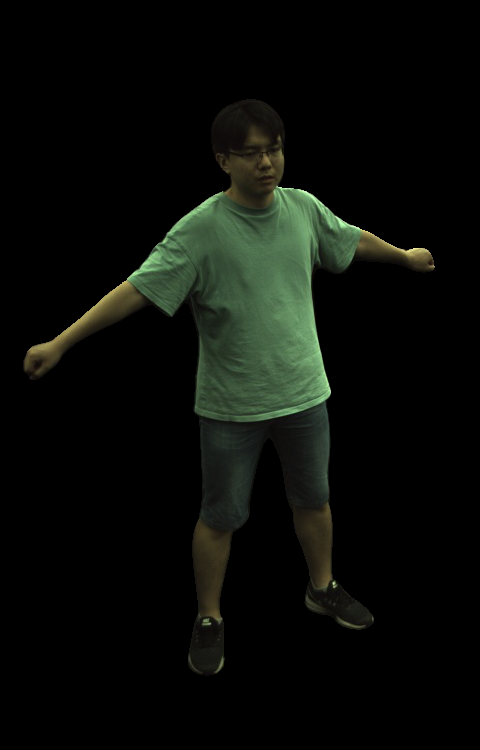} }}
    \vspace{0.5em} %
    \subfloat[\centering]{{\includegraphics[width=0.45\linewidth]{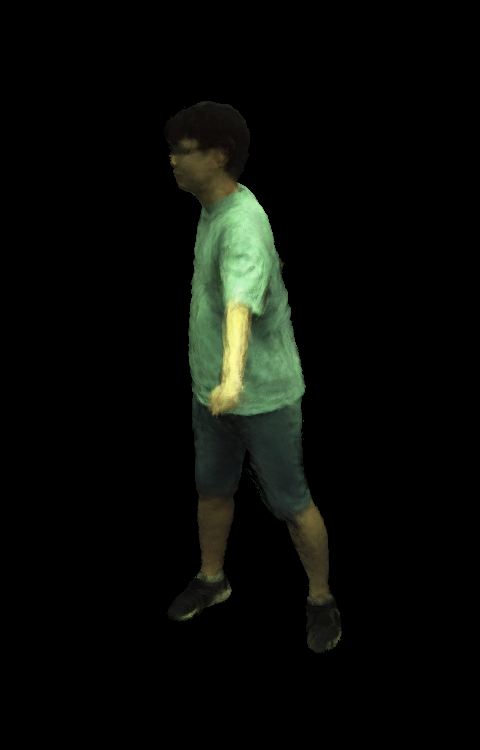} }}%
    \subfloat[\centering]{{\includegraphics[width=0.45\linewidth]{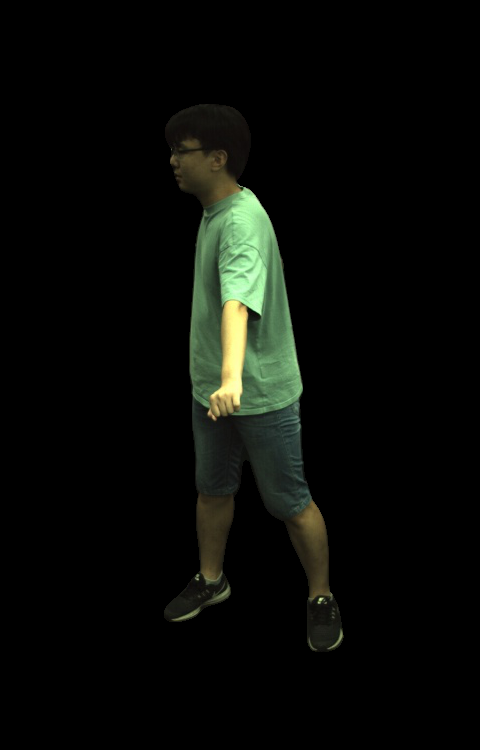} }}%
    \caption{Qualitative results on ZJU-MoCap data. Comparison between (a) novel view synthesis and (b) ground truth, and (c) novel pose synthesis and (d) ground truth.}%
    \label{fig:zju_images}%
\end{figure}

We also show that our approach works on alternative data by training the model on a short sequence of the ZJU-MoCap dataset. 
The quality of synthesized frames is comparable to those reported in \cite{sanerf}, although a comparison is difficult as we employed a distinct training configuration. As a consequence, we present qualitative outcomes only to validate the overall efficacy of our model on foreign data in Figure \ref{fig:zju_images}.

We trained another model on the hand wave sequence depicted in Figure \ref{fig:teaser} with only 5 selected frames of the 85 frame sequence. The frames were chosen in order to sample the whole range of the executed motion to obtain the best result in synthesizing unseen poses. The full result of the reconstructed 85 frames in a novel view can be seen as a video in the supplementary materials, showcasing the possibility to render natural looking human gestures with our approach. 
Notice the qualitative difference of the results when optimizing the SMPL parameters during training. This results in a more accurate representation, leading to sharper results and mitigation of holes in the arms, which were not fitted accurately. Furthermore, it is evident, that the model produces artifacts at the right hand. Since the training dataset consistently depicts the hand resting on the leg, the model faces difficulty in differentiating the distinguishing characteristics between the hand and the leg. Consequently, errors occur when the hand is in motion. This limitation can be mitigated by adding according training samples.

\subsection{Ablation Study}

To show the influence of the main components of the network, we present ablation results in Table \ref{Tab:ablation}. For this study, we used the dataset with the 20 sampled frames from our arm rotation sequence as in the novel pose and view synthesis results. We show the results of the full approach, and an ablation of the crucial components that allow for pose dependent feature learning. First, the replacement of the NeRF ResNet with the conventional NeRF MLP, without having the encoded pose as an input. Furthermore, we show the results without the \textit{uvh} remapping module, and without the ResNet and the remapping module. For the sake of comparison, we trained all models for 100,000 steps in this study.
As expected, we got the best results for the novel pose and the novel view synthesis with our full approach. In the novel view synthesis, it appears that one of the pose dependent modules is sufficient to nearly reach the quality of the full approach. Unexpectedly, in the novel pose synthesis study, the network performed better with no extra modules compared to only the remapping module. It's not clear if this result is expressive about the quality of the produced images, as the largest influence on those values is the shifted appearance. This greatly reduces the PSNR values, even though the results might visually look natural and as expected.

\begin{table}[ht]
\centering
\setlength\tabcolsep{0pt}
\begin{tabular*}{\linewidth}[t]{@{\extracolsep{\fill}} lll }
\hline
& Novel view & Novel pose \\
\hline
\noalign{\vspace{2pt}} 
full & \textbf{28.64} & \textbf{19.16} \\
w/o ResNet & 28.24 & 18.52 \\
w/o Remapping & 28.20 & 19.04 \\
w/o Both & 26.82 & 18.98 \\
\hline
\end{tabular*}
\caption{Quantitative analysis of the ablation study: Evaluation of Peak Signal-to-Noise Ratio (PSNR) in decibel (dB) inside the non-masked area of the ground truth images. The best results are highlighted in bold font.}
\label{Tab:ablation}
\end{table}

\section{Conclusions and Future Research}

We have introduced a novel NeRF-based framework for pose-dependent rendering of human performances. Our approach involves warping the NeRF space around an SMPL body mesh, resulting in a surface-aligned representation that can be animated through skeletal joint parameters. These joint parameters serve as additional input to the NeRF, allowing for pose-dependent appearances in the rendered output. Moreover, our representation includes 2D UV coordinates on the mesh texture map and the distance between the query point and the mesh, facilitating efficient learning despite mapping ambiguities and random visual variations.

Through extensive experiments, we have demonstrated the effectiveness of our approach in generating high-quality renderings for novel-view and novel-pose synthesis. The results showcase the capabilities of our framework in producing visually appealing and controllable 3D human models, offering realistic and pose-dependent renderings, even from a very limited set of training frames.
In future, we plan to train the model on a more diverse set of poses to enable even more accurate reconstructions.

Although our model produces high-fidelity renderings, there remain aspects and challenges that have room for enhancement.
Primarily, as in similar methods, the model struggles with rendering individual fingers, as visible in Figure \ref{fig:teaser}. We have been using the standard SMPL model, which does not capture hand geometry well and will improve by using models specialized on hands, such as SMPL-H, as well as custom model fitting for finger parameters.

\section*{\uppercase{Acknowledgements}}

This work has partly been funded by the German Federal Ministry of Education and Research (Voluprof, grant no. 16SV8705) and the German Federal Ministry for Economic Affairs and Climate Action (ToHyVe, grant no. 01MT22002A).

\bibliographystyle{apalike}
{\small
\bibliography{example}}

\end{document}